\documentclass[letterpaper,twocolumn,fleqn]{article} 

\usepackage{ist}
\usepackage{cite}
\usepackage{mathtools}
\usepackage{titlesec}
\usepackage{amsmath}
\usepackage{amssymb}
\usepackage[]{algorithm2e}
\usepackage{changepage}
\usepackage{url}
\usepackage{multicol}
\usepackage{xcolor}
\usepackage{listings}
\usepackage{subfigure}
\usepackage{comment}
\usepackage[justification=centering]{caption}

\title{Emotion Recognition Using Convolutional Neural Networks}

\author{Shaoyuan Xu\textsuperscript{ a}, Yang Cheng\textsuperscript{ a}, Qian Lin\textsuperscript{ b}, Jan Allebach\textsuperscript{ a}\\\textsuperscript{ a}School of Electrical and Computer Engineering, Purdue University, West Lafayette, IN 47906, U.S.A.\\\textsuperscript{b}HP Labs, Palo Alto, CA 94304, U.S.A.}

\begin{document}
\maketitle 

\thispagestyle{empty} 


\titlespacing*{\chapter} {0pt}{50pt}{40pt}
\titlespacing*{\section} {0pt}{3.5ex plus 1ex minus .2ex}{2.3ex plus .2ex}
\titlespacing*{\subsection} {0pt}{3.25ex plus 1ex minus .2ex}{1.5ex plus .2ex}

\begin{abstract}
Emotion has an important role in daily life, as it helps people better communicate with and understand each other more efficiently. Facial expressions can be classified into 7 categories: angry, disgust, fear, happy, neutral, sad and surprise. How to detect and recognize these seven emotions has become a popular topic in the past decade. In this paper, we develop an emotion recognition system that can apply emotion recognition on both still images and real-time videos by using deep learning.

We build our own emotion recognition classification and regression system from scratch, which includes dataset collection, data preprocessing , model training and testing. Given a certain image or a real-time video, our system is able to show the classification and regression results for all of the 7 emotions. The proposed system is tested on 2 different datasets, and achieved an accuracy of over 80\%. Moreover, the result obtained from real-time testing proves the feasibility of implementing convolutional neural networks in real time to detect emotions accurately and efficiently.

\end{abstract}

\section{1 Introduction}

As one of the most important features of human beings, emotion helps people communicate with and understand each other more efficiently. Therefore, detecting and recognizing various emotions has always been a popular topic. People detect emotions through different methods, such as voice intonation, body language and even electroencephalography \cite{AbhangEmotionRecognition}. However, the most intuitive and practical way of detecting and recognizing emotions is still through facial expressions. In this paper, we propose a system to detect emotions by examining facial expressions. In our system, we follow the research work proposed by Paul Ekman \cite{EkmanUniversals}, where the emotions are categorized into 7 classes: angry, disgust, fear, happy,neutral, sad and surprise, except that the category neutral is replaced with contempt.

There has been a lot of research work on emotion recognition, most of which uses traditional computer vision methods, such as LBP \cite{ShanFacialExpression}; and machine learning classification methods, such as SVM \cite{ShaoyuanSVM}. However, satisfying results could not be achieved due to the limitations of these methods, such as inadaptability to the change of facial muscles. Therefore, we have put much effort in investigating a new approach that take advantages of deep learning\footnote{Research supported by HP Labs, Palo Alto, CA 94304.}.

\begin{figure*}[htb]
     \centering 
     \setlength{\abovecaptionskip}{0.2cm}
      \includegraphics[width=2.1\columnwidth]{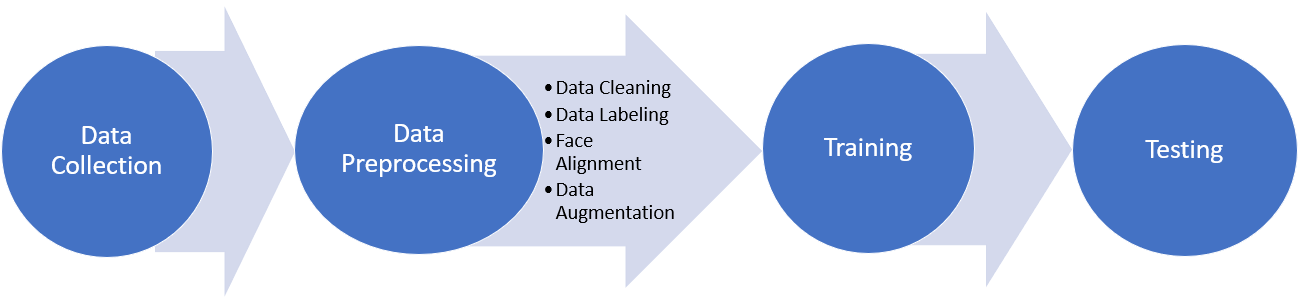}
    \caption{Flowchart of Emotion Recognition Project.}
    \label{EmotionRecognitionFlowchart}
  \end{figure*}

There is also substantial research work done on emotion recognition using deep learning such as traditional model training methods using a specific network \cite{ERDL_cite}, or combining deep learning with machine learning such as LBP \cite{ERLBP_cite}. Although they obtain comparably high accuracy, there are two aspects that need to be improved. Firstly, most of them use traditional network structures such as VGG Net, Alex Net, or Google Net (including the improved versions of these network structures). This results in a large model size; so that it is extremely difficult to do real time emotion recognition. Secondly, most of the proposed systems only consider the classification scenario where the intensity information is missing in the results. But in practical usage, intensity information is as important as the classification result, because we want to know not only what emotions people have, but also the level of those emotions. In this paper, we solve these two problems by selecting an appropriate network structure for an accurate real-time emotion recognition. At the same time, we extend our classification results to the regression scenario so that the intensity information can be concluded from the results. We trained our emotion recognition model for both the classification scenario and the regression scenario. Figure \ref{EmotionRecognitionFlowchart} shows the flowchart of our emotion recognition project.

The paper is organized as follows. In the rest of Section 1, we introduce the seven classes of emotions and an overview of our approaches. Sections 2 and 3 describe how our emotion recognition system is trained using two different models, a classification model and a regression model. The conclusion is provided in Section 4.
  
\section{2 Seven Classes of Emotions}

\begin{figure}[ht]
     \centering 
     \setlength{\abovecaptionskip}{0.2cm}
      \includegraphics[width=1.0\columnwidth]{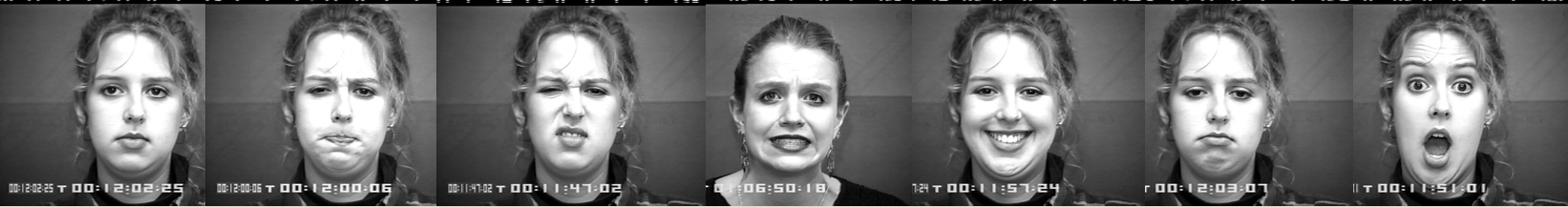}
    \caption{Seven universal emotions: Neutral, Angry, Disgust, Fear, Happy, Sad and Surprise.}
    \label{SevenEmotions}
  \end{figure}

There are seven universal emotions: angry, disgust, fear, happy, neutral, sad and surprise. Examples of those emotions are shown in Figure \ref{SevenEmotions} \cite{CK+1} \cite{CK+2} and each emotion is described by some characteristics \cite{HumintellEmotions}.

\begin{description}
  \item[$\bullet$]Angry: eyebrows are pulled down, upper eyelids are pulled up, lower eyelids are pulled up, margins of lips are rolled in and lips may be tightened.

  \item[$\bullet$]Disgust: eyebrows are pulled down, nose is wrinkled and upper lips are pulled up and loose.

  \item[$\bullet$]Fear: eyebrows are pulled up, upper eyelids are pulled up and mouth is stretched.

  \item[$\bullet$]Happy: muscles around the eyes are tightened, ``Crows Feet'' wrinkles appear around eyes, cheeks are raised and lip corners are raised diagonally.

  \item[$\bullet$]Sad: inner corners of eyebrows are raised, eyelids are loose and lip corners are pulled down.

  \item[$\bullet$]Surprise: entire eyebrows are pulled up, eyelids are pulled up and mouth hangs open.
\end{description}

\section{3 Emotion Recognition Classification Training}
\subsection{Data Collection}

\begin{figure}[ht]
     \centering 
     \setlength{\abovecaptionskip}{0.2cm}
      \includegraphics[width=1.0\columnwidth]{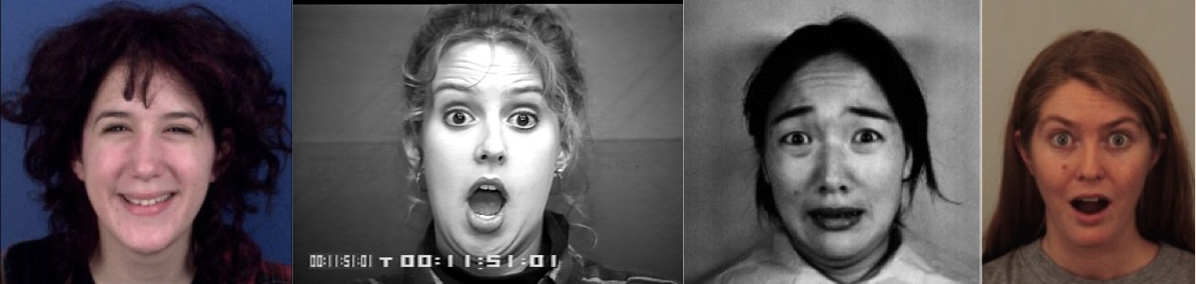}
    \caption{Sample images of MUG-FED, CK+, Jaffe, and KDEF.}
    \label{SampleTrainingImages}
  \end{figure}

Due to the lack of public datasets for emotion recognition tasks and the low quality of existing datasets, collecting enough datasets and examining them becomes the first challenging task. Firstly, there are 4 publicly available datasets: MUG-FED \cite{MUG}, CK+ \cite{CK+1} \cite{CK+2}, Japanese Female Facial Expression (Jaffe) \cite{Jaffe} and KDEF \cite{KDEF}. Figure \ref{SampleTrainingImages} shows some sample images from these four datasets, and Table \ref{TrainingImagesStatistics} contains the statistics. Secondly, we have collected our own dataset for testing.

\begin{table*}[htb]
\small
\centering
\renewcommand\arraystretch{1.5}
\setlength{\abovecaptionskip}{5pt}
\label{my-label}
\caption{Statistics of the 4 collected datasets.}
\begin{tabular}{| p{0.20\columnwidth} | p{0.30\columnwidth} | p{0.20\columnwidth} | p{0.30\columnwidth} | p{0.18\columnwidth}| p{0.17\columnwidth} | p{0.20\columnwidth} | p{0.20\columnwidth} |}
\hline
Database  & Facial Expression & $\#$ of Subjects & $\#$ of Images & Gray/Color & Size (pixel) & Ground Truth & Type  \\ \hline
Extended Cohn-Kanade Dataset (CK+) & Angry, Contempt, Disgust, Fear, Happy, Neutral, Sad and Surprise & 123 & 593 image sequences (327 sequences having discrete emotion labels) & Mostly gray & 640 $\times$ 490 & Facial expression labels and FACS labels & Posed; spontaneous smiles    \\ \hline
Japanese Female Facial Expression (Jaffe) & Angry, Disgust, Fear, Happy, Neutral, Sad and Surprise & 10 & 213 static images & Gray & 256 $\times$ 256 & Facial expression labels & Posed \\ \hline
Multimedia Understanding Group (MUG-FED) & Angry, Disgust, Fear, Happy, Neutral, Sad and Surprise & 86 & 1462 sequences with more than 100K images & Color & 896 $\times$ 896 & Facial expression and land mark (LM) labels & Posed \\ \hline
The Karolinska Directed Emotional Faces (KDEF) & Angry, Disgust, Fear, Happy, Neutral, Sad and Surprise & 70 & 4900 images & Color & 562 $\times$ 762 & Facial expression labels & Posed \\ \hline
\end{tabular}
\label{TrainingImagesStatistics}
\end{table*}

\subsection{Data Preprocessing}
\subsubsection{Dataset Cleaning}

Since most of the public datasets contain raw images, very few of them can be directly used without further examination. Therefore, these datasets need to be cleaned in the first place. 

There are 52 subjects in the MUG-FED Dataset. For each subject, it has 5-7 emotions and for each emotion, it has 3-7 attempts. And since all of the images are video frames, each emotion starts from neutral to the emotional expression of the strongest intensity and returns to neutral. Therefore, only the images that contain facial expressions of strong intensity should be selected. Table \ref{MUG-FED} shows the statistics of the MUG-FED Dataset after it is cleaned. It also provides us with 161 manually labeled images, which is used for validation. Besides these 4 datasets obtained from online sources, an additional 490 images were collected by us and are used to validate the model.

\begin{table*}[htb]
\small
\centering
\renewcommand\arraystretch{1.5}
\setlength{\abovecaptionskip}{5pt}
\setlength{\belowcaptionskip}{0.1cm}
\label{my-label}
\caption{Statistics of the MUG-FED Dataset after dataset cleaning up.}
\begin{tabular}{| p{0.21\columnwidth} | p{0.11\columnwidth} | p{0.13\columnwidth} | p{0.11\columnwidth} | p{0.11\columnwidth}| p{0.13\columnwidth} | p{0.11\columnwidth} | p{0.14\columnwidth} | p{0.11\columnwidth} |}
\hline
Emotion Type  & Angry & Disgust & Fear & Happy & Neutral & Sad & Surprise & Total \\ \hline
$\#$ of Images & 6220 & 4856 & 4605 & 9329 & 3719 & 5562 & 5623 & 39914   \\ \hline
\end{tabular}
\label{MUG-FED}
\end{table*}

Eventually, there are 3 datasets for training and 3 datasets for validation, as shown in Table \ref{TrainingValidation}.

\begin{table}[htb]
\small
\centering
\renewcommand\arraystretch{1.5}
\setlength{\abovecaptionskip}{5pt}
\label{my-label}
\caption{Training and Validation datasets for classification training.}
\begin{tabular}{| p{0.18\columnwidth} | p{0.17\columnwidth} | p{0.5\columnwidth} |}
\hline
Dataset  & Training & Validation\\ \hline
Name & MUG-FED, CK+ and Jaffe & MUG-FED (Manually Labeled by author), KDEF and Images of myself \\ \hline
$\#$ of Images  & 41029 & 1867\\ \hline
\end{tabular}
\label{TrainingValidation}
\end{table}

\subsubsection{Face Alignment}

Face alignment is another key step in dataset pre-processing. The purpose is to remove potential uncertainties when applying our emotion recognition approach to real-time videos. For example, the position and the angle of the subject's head are changing as the video plays, which could affect the accuracy of the classification results if the face is not aligned in advance. With face alignment, the position of the head is aligned and the scale of the head is adjusted to have the same size, which eliminates the influence of any existing distortions on the recognition results. 

We propose a novel face alignment algorithm that shows superior results compared to any existing method. Firstly, a face detector is used to detect the face in an image, then the Land Mark (LM) detector \cite{LMDetector} is used to detect 68 landmark points of the face. A rotation matrix is then obtained based only the eye center coordinates. The traditional method uses the rotation matrix for face alignment. However, the resultant images can contain comparably useless background of large area and the eyes in different images are not at the same horizontal level, resulting in unsatisfying face classification results.

\begin{figure}[ht]
     \centering 
     \setlength{\abovecaptionskip}{0.2cm}
      \includegraphics[width=1.0\columnwidth]{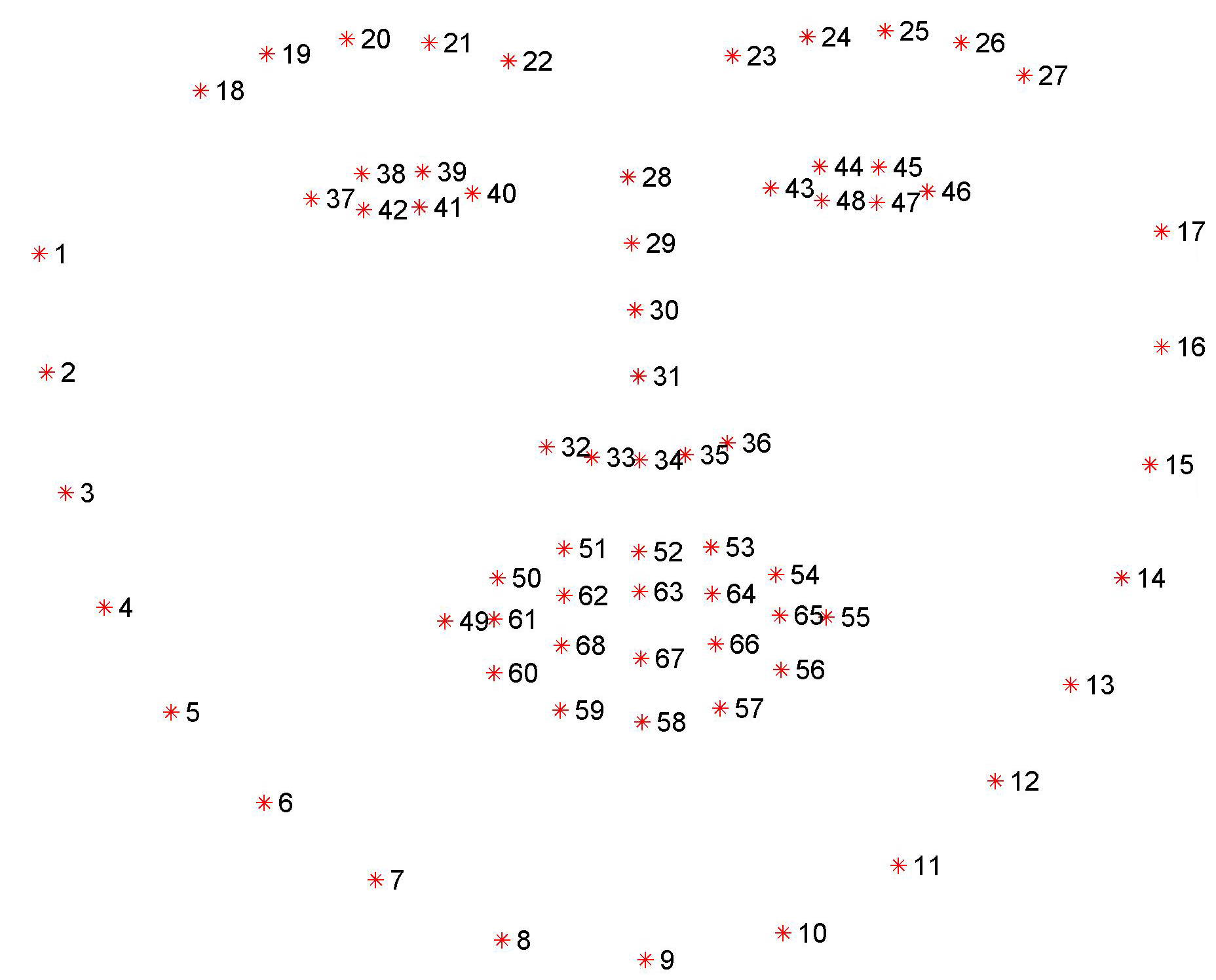}
    \caption{68 points facial landmark system.}
    \label{figure_68_markup}
  \end{figure}

\begin{figure}[ht]
   \centering
    \subfigure[]{
      \includegraphics[width=0.2\columnwidth]{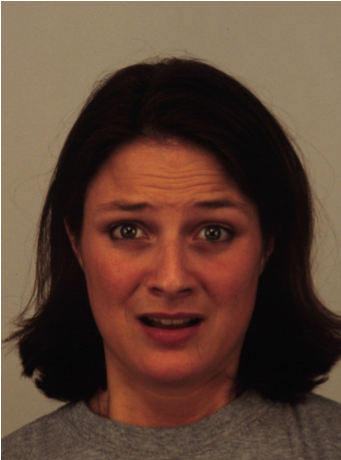}
      \label{KDEF1}
    }
    \subfigure[]{
      \includegraphics[width=0.2\columnwidth]{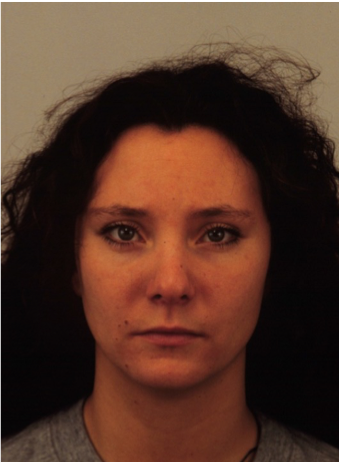}
      \label{KDEF2}
    }
    \subfigure[]{
      \includegraphics[width=0.2\columnwidth]{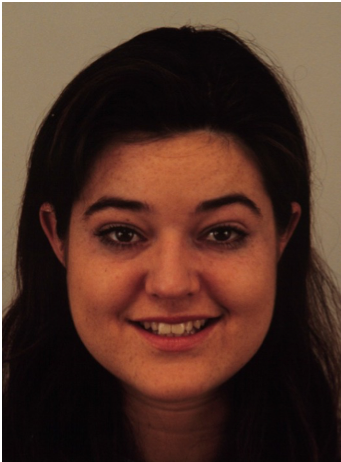}
      \label{KDEF3}
    }
    \\
    \subfigure[]{
      \includegraphics[width=0.2\columnwidth]{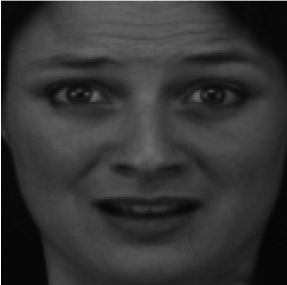}
      \label{KDEF11}
    }
    \subfigure[]{
      \includegraphics[width=0.2\columnwidth]{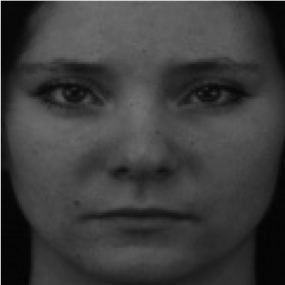}
      \label{KDEF21}
    }
    \subfigure[]{
      \includegraphics[width=0.2\columnwidth]{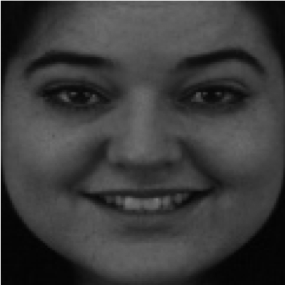}
      \label{KDEF31}
    }
    \caption{Comparison of images before and after face alignment.}
    \label{Facealignment}
  \end{figure}

We improve the traditional face alignment by adding one more step. The aforementioned rotation matrix gets the coordinates of the 68 landmark points in the new coordinate system. Then we use the $1^{st}$, $9^{th}$ and $17^{th}$ landmark points to get the left, bottom and right boundary of the face. The definition and the location of these facial landmark points are shown in Figure \ref{figure_68_markup} \cite{300wdatabase} \cite{semiautomaticLM} \cite{300wchallenge}. To get the top boundary, we stipulate that the length from the top boundary to the eye center is one-third of the height of the image. We use the boundary information to crop the original image into the one that contains smaller margins. Finally, the eyes of different images are adjusted to be at the same horizontal level. Figure \ref{Facealignment} shows some sample images before and after face alignment. Note that all the images after face alignment are re-scaled to $128 \times 128$.

\subsubsection{Data Augmentation}

To increase the robustness of our model and to prevent it from being over-fitted, we apply data augmentation on the training dataset after face alignment. For each image, 7 images of different brightness and 28 images of different degree of blurring are created, resulting in a final training set with 1,148,812 images. Table \ref{DataAugmentation} shows the statistics of the data augmentation.

\begin{table*}[htb]
\small
\centering
\renewcommand\arraystretch{1.5}
\setlength{\abovecaptionskip}{5pt}
\label{my-label}
\caption{Data augmentation of classification training dataset.}
\begin{tabular}{|p{0.30\columnwidth} | p{0.17\columnwidth} | p{0.3\columnwidth} | p{0.18\columnwidth} | p{0.30\columnwidth} |}
\hline
$\#$ of Images Before Face Alignment & Brightness Change & Blur (Gaussian, Average, Median) & Total Multiples & $\#$ of Images After Face Alignment\\ \hline
41029 & 7 $\times$ & 4 $\times$ & 28 $\times$ & 1148812 \\ \hline
\end{tabular}
\label{DataAugmentation}
\end{table*}

\subsection{Model Training}

In order to apply our emotion recognition system to real-time video, the model needs to be comparably small in size and fast in speed. We have tested several pre-traind models, such as the VGG-S \cite{VGG-S}, on real-time video with multiple rounds of fine-tuning. The validation accuracies are less than 60\% which is far below our requirements. Moreover, the size of the VGG-S model is more than 500 MB which is too large to be implemented efficiently.

  \begin{figure*}[ht]
     \centering 
     \setlength{\abovecaptionskip}{0.2cm}
     \setlength{\belowcaptionskip}{0.2cm}
      \includegraphics[width=2.1\columnwidth]{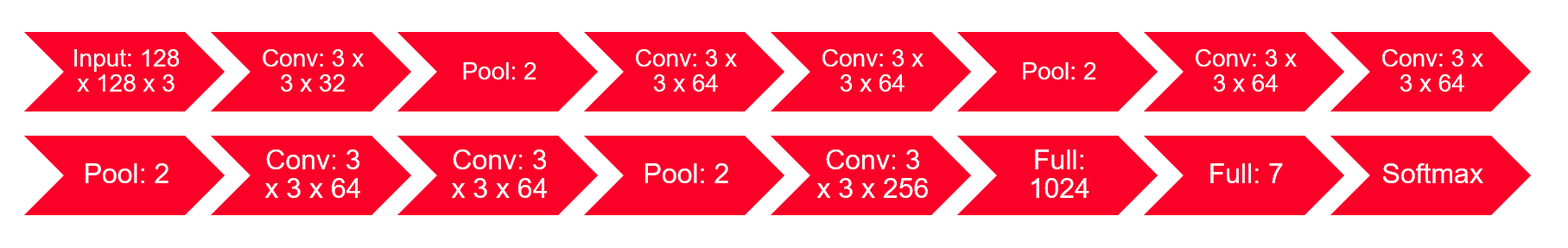}
    \caption{Framework of the classification model.}
    \label{ClassificationModel}
  \end{figure*}
  
To reduce the model size, we modified the original VGG-S  \cite{Model} model by reducing the kernel size and channel number as shown in Figure \ref{ClassificationModel} \cite{LMDetector}. Compared to the original VGG-S model, our model has a size of only 12.1 MB. It takes only 4.5 hours to train on more than 1 million images for 50,000 iterations. Besides a smaller size of the model, the validation accuracy obtained by using the new model reaches 85\%, which is significantly higher than our previous results. The reason of getting higher accuracy with a model of smaller size is that, if we want to train with the original large-size VGG-S model, it requires millions of raw images and weeks of time to train from scratch which is impossible. Which means, with our training set, smaller model will have higher accuracy. And the modified model can be trained faster and more efficiently with much smaller data set and much less time, in our case, 40k images and 4.5 hours. 
  
\section{4 Emotion Recognition Regression Training}
\subsection{Introduction}

Although our emotion recognition classification model works well, it has its own drawback. And it is especially obvious when the classification model is applied in a real-time demo. 

Since our classification training dataset includes a lot of emotions that are not obvious or are of low intensity, this making them similar to one of the emotion categories in particular: neutral. This causes the prediction on the real-time video to be jittery, since in most cases, for example, a person does not need to express his or her happiness by a drop-jaw smile. And also, given an image or a frame of the video, our classification model can only tell if the facial expression is angry, disgust, fear, happy, neutral, sad or surprise; in other words, it is not able to tell the intensity of the emotion. 

To solve these two problems, a regression model is used, where the ground truth labels become the intensities of the emotion, such as 20\% happy and 80\% neutral or 40\% sad and 60\% neutral. This additional information about the intensity of the emotion can be useful, especially in real-time videos.

\subsection{Data Collection}

Among the four datasets collected from the online sources, only the MUG-FED Dataset is used because of the large number of images the dataset includes. However, the MUG-FED Dataset is more like an ``in-the-lab'' dataset, where all of the emotions included are standardized and all the images have the same background and consist of a purely frontal face. Since there are very few public in-the-wild datasets, especially for the task of emotion recognition, we collected our own dataset to train the model on a more ``in-the-wild'' dataset.

Until now, we have collected more than 7000 ``in-the-wild'' images containing facial expressions and we name this dataset as Emotion Intensity in the Wild Dataset. Table \ref{EmotinoIntheWild} shows the statistics of this dataset. It is worth noting that this dataset includes images containing heads at different angles, people with different races and ages, and backgrounds of different lighting conditions.

\begin{table*}[htb]
\small
\centering
\renewcommand\arraystretch{1.5}
\setlength{\abovecaptionskip}{5pt}
\label{my-label}
\caption{Statistics for Emotion Intensity In the Wild Dataset.}
\begin{tabular}{| p{0.11\columnwidth} | p{0.10\columnwidth} | p{0.13\columnwidth} | p{0.08\columnwidth} | p{0.11\columnwidth} | p{0.13\columnwidth} | p{0.07\columnwidth} | p{0.15\columnwidth} | p{0.5\columnwidth} |}
\hline
Dataset & Angry & Disgust & Fear & Happy & Neutral & Sad & Surprise & Total (Without Neutral) \\ \hline
20\% & 194 & 147 & 93 & 236 & 1093 & 210 & 91 & 971 \\ \hline
40\% & 323 & 218 & 103 & 230 &  & 164 & 218 & 1256 \\ \hline
60\% & 221 & 243 & 134 & 320 &  & 137 & 279 & 1334 \\ \hline
80\% & 207 & 198 & 151 & 295 &  & 81 & 316 & 1248 \\ \hline
100\% & 227 & 178 & 157 & 286 &  & 84 & 420 & 1352 \\ \hline
Total & 1172 & 984 & 638 & 1367 & 1093 & 676 & 1324 & 6161
(7254 Total With Neutral) \\ \hline
\end{tabular}
\label{EmotinoIntheWild}
\end{table*}

\begin{table*}[!htbp]
\small
\centering
\renewcommand\arraystretch{1.5}
\setlength{\abovecaptionskip}{5pt}
\setlength{\belowcaptionskip}{0.1cm}
\label{my-label}
\caption{Regression training results.}
\begin{tabular}{| p{0.5\columnwidth} | p{0.24\columnwidth} | p{0.27\columnwidth} | p{0.32\columnwidth} | p{0.40\columnwidth} |}
\hline
Dataset & Training Loss & Validation Loss & Regression RMSE & Classification Accuracy \\ \hline
Emotion Intensity In the Wild & 0.13 & 0.3 & 0.129 & 77.2\% \\ \hline
Combined Dataset & 0.122 & 0.239 & 0.123 & 76\% \\ \hline
\end{tabular}
\label{RegressionTrainingResults}
\end{table*}

To train the regression model, we use both the MUG-FED and Emotion Intensity In the Wild datasets.

\subsection{Data Preprocessing}
\subsubsection{Dataset Cleaning}

\begin{figure*}[ht]
     \centering 
     \setlength{\abovecaptionskip}{0.2cm}
      \includegraphics[width=2.1\columnwidth]{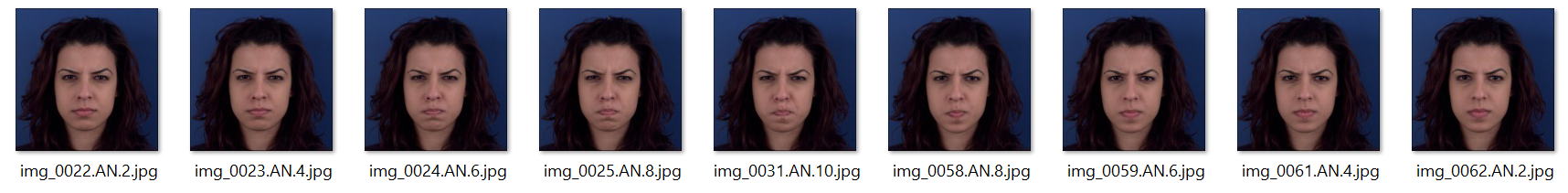}
    \caption{Sample regression images from MUG-FED Dataset. These images are from one of the attempts that a subject does which have the intensities from 20\% (neutral) to 100\%, and back to 20\% in steps of 20\%. Noting that the numbers: 2, 4, ... after the emotion label $AN$ are the intensity labels, $2$ corresponds to $20\%$ etc.}
    \label{MUGRegressionSample}
  \end{figure*}

Before an existing dataset obtained from online sources is used, it needs to be examined in the first place. As we introduced in the previous section, the images of the MUG-FED Dataset are consecutive frames obtained from videos. In a video for a specific emotion, the emotion starts from neutral to 100\% facial expression and gradually returns to neutral. Therefore, for each attempt of expressing emotion, we select 9 images that contain facial expression intensities from 20\% to 100\% and back to 20\% with an interval of 20\%. An example is shown in Figure \ref{MUGRegressionSample}. After dataset cleaning, we have collected 7,451 images for training and 981 images for validation for the MUG-FED Dataset. Each of these 7451 images is labeled with the intensity of the emotion.

And for the Emotion Intensity In the Wild Dataset, after excluding some inappropriate images, we have 6141 images for training and 682 images for validation.

\subsubsection{Face Alignment}

The procedure of face alignment is the same as the one introduced in the previous section. We utilized the $1^{st}$, $9^{th}$ and $17^{th}$ landmark points to get the boundary of the faces, cropped them, aligned them and rescaled them to $128 \times 128$.

\subsubsection{Data Augmentation}

We experimented with two strategies of training, one with only the Emotion Intensity In The Wild Dataset, another with the combined dataset of Emotion Intensity In The Wild Dataset and the MUG-FED Dataset. The method of data augmentation remains the same, which includes changing brightness and blurring the images and gives the final training dataset, contains 6,141 images before data augmentation and 171,948 images after, and final validation dataset, containing 682 images before data augmentation and 19,096 images after for Emotion Intensity In The Wild Dataset. For the combined dataset, the final training dataset, containing 13,592 images before data augmentation and 380,576 images after, and the final validation dataset, containing 1,663 images before data augmentation and 46,564 images after, as shown in Table \ref{RegressionTraining}.


\begin{table}[htb]
\small
\centering
\renewcommand\arraystretch{1.5}
\setlength{\abovecaptionskip}{5pt}
\label{my-label}
\caption{Dataset statistics for regression training.}
\begin{tabular}{| p{0.48\columnwidth} | p{0.18\columnwidth} | p{0.18\columnwidth} |}
\hline
Dataset & Training & Validation \\ \hline
Emotion Intensity In the Wild (After data augmentation)& 6141 (171948) & 682 (19096) \\ \hline
Combined Dataset \quad \qquad \qquad(After data augmentation)& 13592 (380576) & 1663 (46564) \\ \hline
\end{tabular}
\label{RegressionTraining}
\end{table}

\subsection{Model Training}

The model framework used for the regression training is the same as for our classification model, except that the softmax loss function is replaced with the sigmoid cross entropy loss function. 
\section{5 Experimental Results}

\subsection{Classification Results}

\begin{figure}[ht]
     \centering 
     \setlength{\abovecaptionskip}{0.2cm}
      \includegraphics[width=0.8\columnwidth]{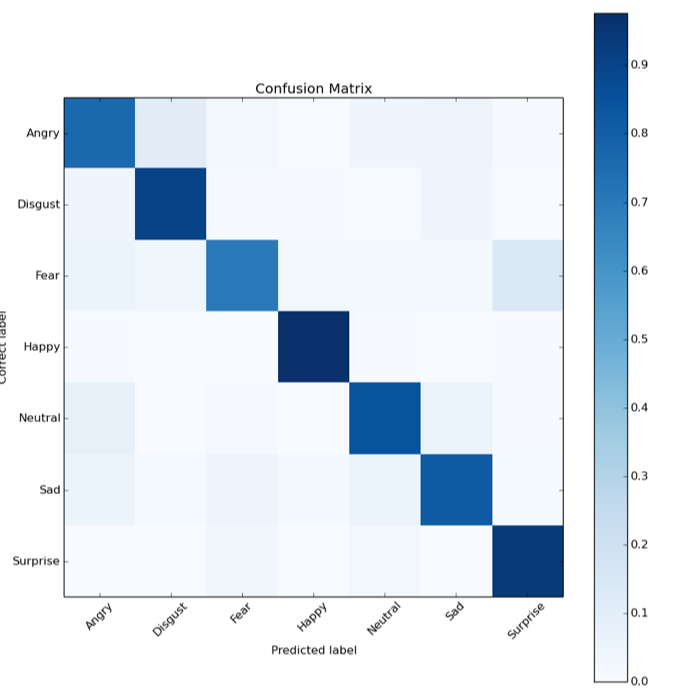}
    \caption{Classification confusion matrix for our classification model validation set. The overall accuracy is 85\%.}
    \label{ConfusionMatrixValidation}
  \end{figure}

As introduced in Section 3, the classification validation accuracy of the classification model is 85\%. Figure \ref{ConfusionMatrixValidation} shows the validation confusion matrix for the validation dataset.

\begin{figure}[ht]
   \centering
    \subfigure[]{
      \includegraphics[width=0.3\columnwidth]{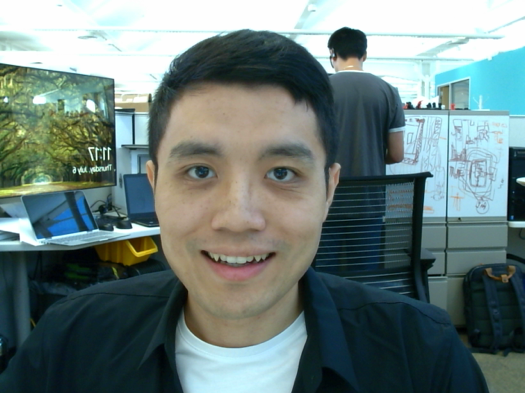}
      \label{SampleTesting1}
    }
    \hfill
    \subfigure[]{
      \includegraphics[width=0.3\columnwidth]{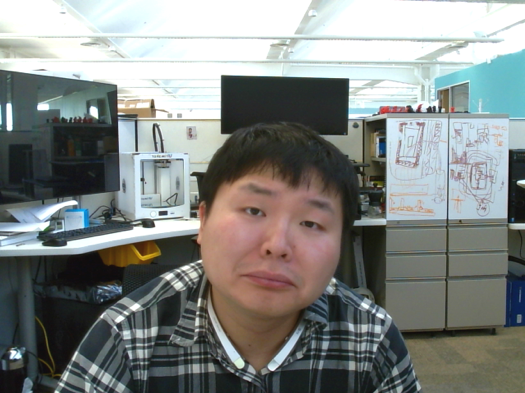}
      \label{SampleTesting2}
    }
    \hfill
    \subfigure[]{
      \includegraphics[width=0.3\columnwidth]{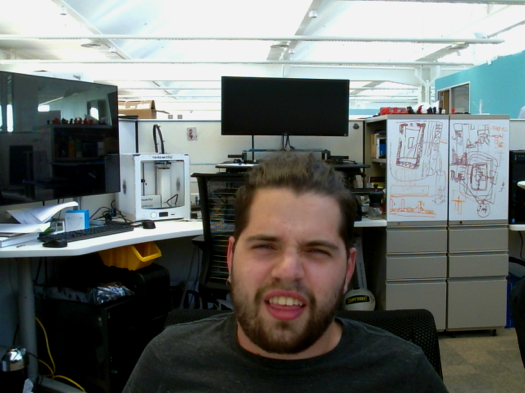}
      \label{SampleTesting3}
    }
    \\
    \subfigure[]{
      \includegraphics[width=0.3\columnwidth]{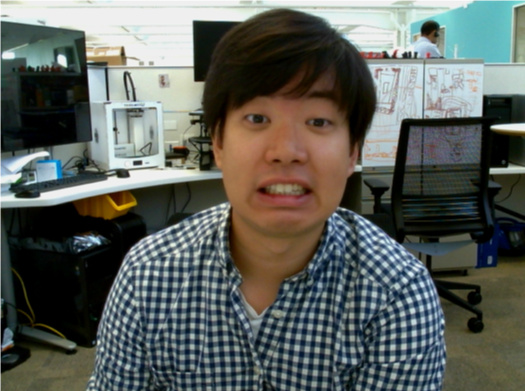}
      \label{SampleTesting4}
    }
    \subfigure[]{
      \includegraphics[width=0.3\columnwidth]{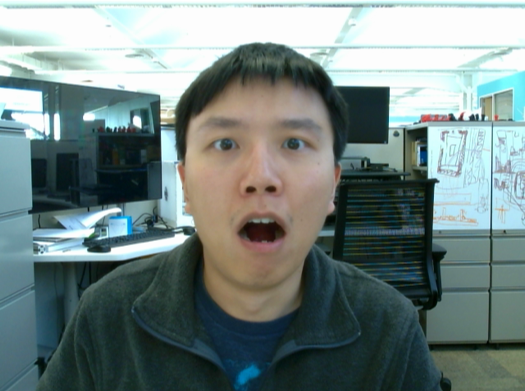}
      \label{SampleTesting5}
    }
    \caption{Sample images from HP Facial Expression Test Set.}
    \label{SampleTestingImages}
  \end{figure}

\begin{figure}[ht]
     \centering 
     \setlength{\abovecaptionskip}{0.2cm}
      \includegraphics[width=0.8\columnwidth]{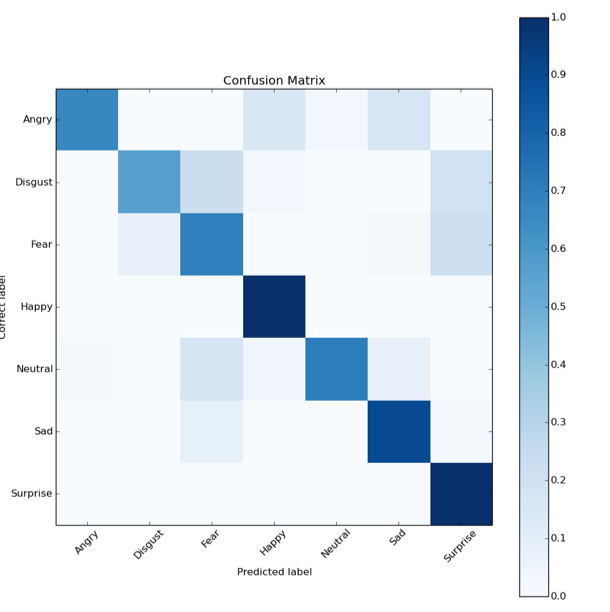}
    \caption{Classification confusion matrix for our self-collected HP Facial Expression Test Set. The overall accuracy is 82\%.}
    \label{ConfusionMatrixHP}
  \end{figure}

In order to test our classification model, we collected our own dataset which is called the HP Facial Expression Test Set, which contains 2443 images. The dataset was collected with 5 subjects doing 7 emotions while being video recorded and the images were selected from the video frames. Our model achieves an accuracy of 82\% and takes only 13.68 seconds to test on the whole testing dataset (0.0056 s/image). This test was conducted on a workstation with an Nvidia Titan X GPU. Figure \ref{SampleTestingImages} shows some sample testing images and Figure \ref{ConfusionMatrixHP} shows the testing confusion matrix.

\subsection{Regression Results}

\begin{figure}[ht]
     \centering 
     \setlength{\abovecaptionskip}{0.2cm}
      \includegraphics[width=0.8\columnwidth]{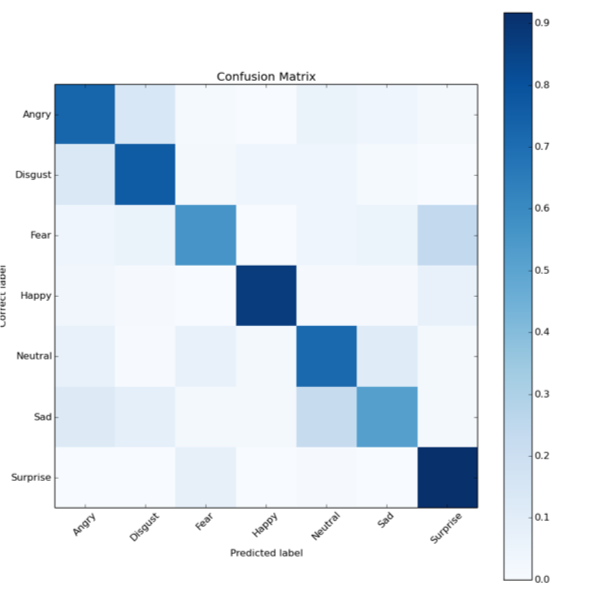}
    \caption{Classification confusion matrix for Emotion Intensity In the Wild Dataset. The overall accuracy is 77.2\%.}
    \label{ConfusionMatrixInthewild}
  \end{figure}

\begin{figure}[ht]
     \centering 
     \setlength{\abovecaptionskip}{0.2cm}
      \includegraphics[width=0.8\columnwidth]{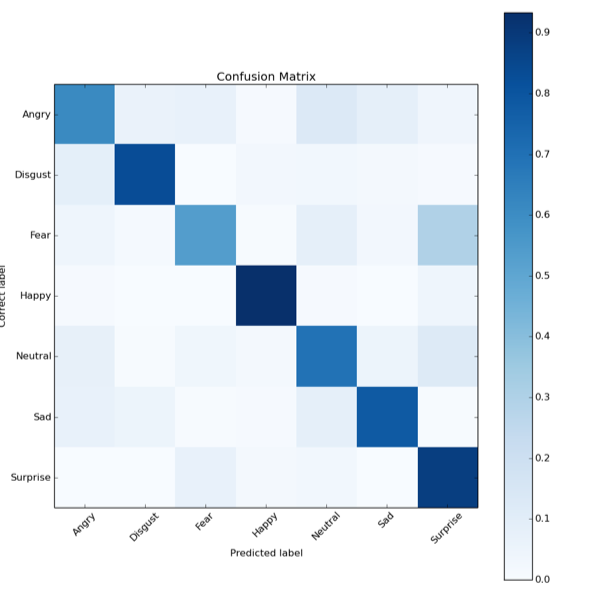}
    \caption{Classification confusion matrix for the Combined Dataset. The overall accuracy is 76\%.}
    \label{ConfusionMatrixCombined}
  \end{figure}

The regression training also gets outstanding results. Figure \ref{ConfusionMatrixInthewild} and Figure \ref{ConfusionMatrixCombined} shows the classification confusion matrices; and Table \ref{RegressionTrainingResults} shows the regression training results. Noting that, for the training and validation loss values, they are sigmoid cross entropy loss, the smaller the better and for the RMSE values, they represent the standard deviation of the prediction errors and are based on the datum that ranges from 0 to 1.
  
As indicated by the high accuracy which is around 77\% and the small Root Mean Squared Error (RMSE) value which is below 0.13, our regression model performs well on the emotion recognition task.
  
\subsection{Real-time Emotion Recognition}

\begin{figure}[htb]
   \centering
    \subfigure[]{
      \includegraphics[width=0.3\columnwidth]{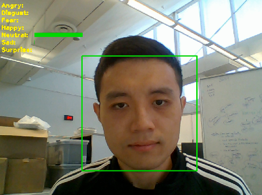}
      \label{SampleRealTime1}
    }
    \hfill
    \subfigure[]{
      \includegraphics[width=0.3\columnwidth]{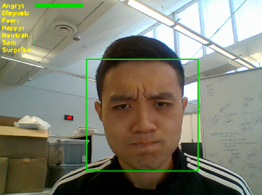}
      \label{SampleRealTime2}
    }
    \hfill
    \subfigure[]{
      \includegraphics[width=0.3\columnwidth]{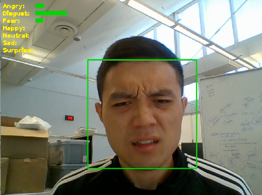}
      \label{SampleRealTime3}
    }
    \\
    \subfigure[]{
      \includegraphics[width=0.3\columnwidth]{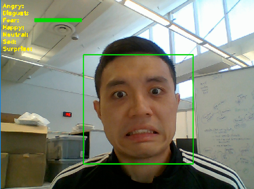}
      \label{SampleRealTime4}
    }
    \subfigure[]{
      \includegraphics[width=0.3\columnwidth]{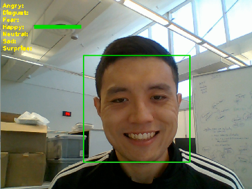}
      \label{SampleRealTime5}
    }
    \subfigure[]{
      \includegraphics[width=0.3\columnwidth]{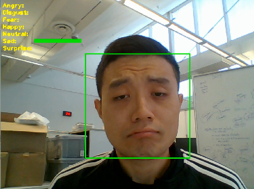}
      \label{SampleRealTime6}
    }
    \subfigure[]{
      \includegraphics[width=0.3\columnwidth]{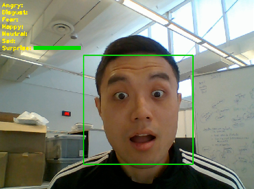}
      \label{SampleRealTime7}
    }
    \caption{Sample result frames from real-time video demo for all seven emotions. The reader should zoom in to be able to see the labeling of the emotion and intensity provided in the upper left corner of each frame.}
    \label{SampleRealTime}
  \end{figure}

Currently, there are not many real-time emotion recognition frameworks, while the existing ones achieve only comparably low accuracy. However, our real-time demo version can detect people's frontal facial expressions accurately. Figure \ref{SampleRealTime} shows some sample results from our real-time emotion recognition demo.

\section{6 Conclusion}

In this paper, we apply emotion recognition using a deep learning method. The whole process includes data collection, data preprocessing, model training and model testing. Our contributions include:

\begin{description}
  \item[$\bullet$] We built an emotion recognition framework from scratch. We first collected four public datasets and manually cleaned them. After that, we implemented data preprocessing, including labeling data, aligning faces and augmenting data. In the training process, we designed our own model based on a VGG-S model but with a much smaller size, better accuracy and improved efficiency.
  \item[$\bullet$] We developed an emotion recognition regression training framework to consider the intensity information of emotions. We collected our own Emotion Intensity In the Wild Dataset and defined a 5-level regression labeling scenario. Our experiment results show that the proposed system can recognize the emotion intensities with promising accuracies.
  \item[$\bullet$] We showed that our model achieved accurate and smooth real-time recognition of both emotion type and intensity by applying it to real-time scenarios.
\end{description}




\small
\bibliographystyle{ieeetr}
\bibliography{cited}

\end{document}